\documentclass[runningheads]{llncs}
\usepackage{booktabs} % table
\usepackage{multirow} % table
\usepackage{graphicx}
\usepackage{algorithm} % algorithm
\usepackage{algorithmic} % algorithm
\usepackage{bm} % bold font
\usepackage{hyperref} % hyper ref
\usepackage {amsfonts,amssymb} 
\usepackage{fancyhdr}
\usepackage{graphicx}
\usepackage{color}

\begin{document}
%%%%%%%%%%%%%%%%
%%%%%%%%%%%%%%%%
%%%%%%%%%%%%%%%%
%%%%%%%%%%%%%%%%
%%%%%%%%%%%%%%%%
\setcounter{page}{100}
\hypersetup{hidelinks}
%%%%%%%%%%%%%%%%
%%%%%%%%%%%%%%%%
%%%%%%%%%%%%%%%%
%%%%%%%%%%%%%%%%
%%%%%%%%%%%%%%%%
\title{Self-Supervised Learning for Building Damage Assessment from Large-scale xBD Satellite Imagery Benchmark Datasets}
\titlerunning{Self-Supervised Learning for BDA from xBD Datasets}
% \author{Zaishuo Xia\inst{1}\and Zelin Li\inst{1}\and Yanbing Bai\inst{1}\thanks{This research was supported by Public Computing Cloud, Renmin University of China.}\and Jinze Yu\inst{2}\and Bruno Adriano\inst{3}}
\author{Zaishuo Xia\inst{1}\and Zelin Li\inst{1}\and Yanbing Bai\inst{1\dagger}\and Jinze Yu\inst{2}\and Bruno Adriano\inst{3}}
\authorrunning{Z.Xia et al.}
\institute{Center for Applied Statistics, School of Statistics, Renmin University of China\and Department of Artificial Intelligent, Connected Robotics Inc., Japan \and RIKEN Center for Advanced Intelligence Project (AIP), Japan \\
$^{\dagger}$ Corresponding author:Yanbing Bai $<$ybbai@ruc.edu.cn$>$}
\maketitle              % typeset the header of the contribution
{
\vspace*{-66mm}\hspace*{-14mm}
\href{https://arxiv.org/abs/2205.15688}{\includegraphics[width=12mm]{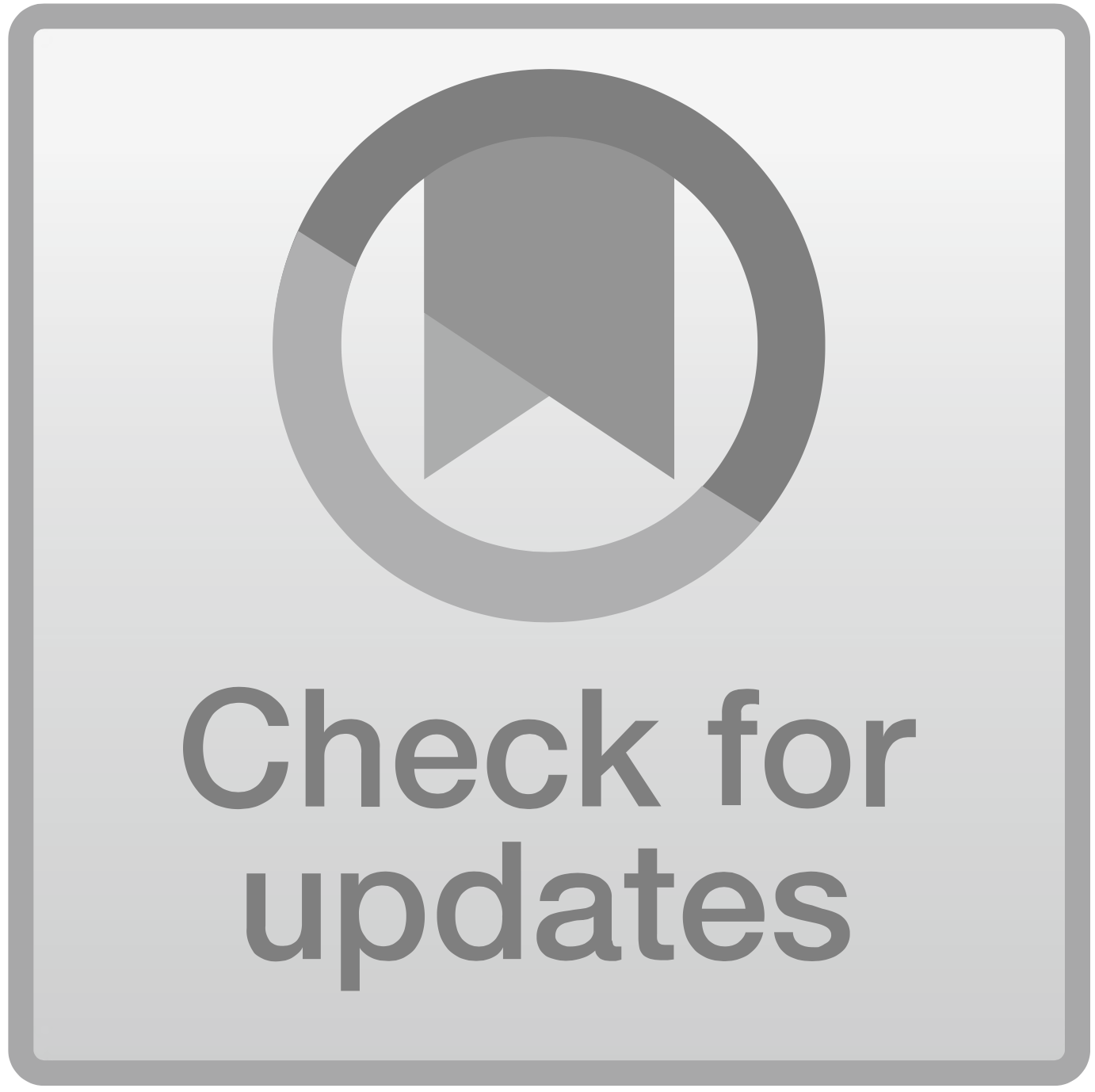}}
\vspace*{46mm}
}
\begin{abstract}
In the field of post-disaster assessment, for timely and accurate rescue and localization after a disaster, people need to know the location of damaged buildings. In deep learning, some scholars have proposed methods to make automatic and highly accurate building damage assessments by remote sensing images, which are proved to be more efficient than assessment by domain experts. However, due to the lack of a large amount of labeled data, these kinds of tasks can suffer from being able to do an accurate assessment, as the efficiency of deep learning models relies highly on labeled data. Although existing semi-supervised and unsupervised studies have made breakthroughs in this area, none of them has completely solved this problem. Therefore, we propose adopting a self-supervised comparative learning approach to address the task without the requirement of labeled data. We constructed a novel asymmetric twin network architecture and tested its performance on the xBD dataset. Experiment results of our model show the improvement compared to baseline and commonly used methods. We also demonstrated the potential of self-supervised methods for building damage recognition awareness.

\keywords{Self-Supervised Learning  \and Building Damage Assessment \and Satellite Imagery \and xBD Dataset}
\end{abstract}

\renewcommand{\thefootnote}{}
\footnote{\noindent\hspace*{-3.5mm} \copyright\ Springer Nature Switzerland AG 2022 \\
 \noindent\hspace*{-3.5mm} C. Strauss et al. (Eds.): DEXA 2022, LNCS 13108, pp. 100–113, 2022.\\
 \noindent\hspace*{-3.5mm} \textcolor{blue}{\url{https://doi.org/10.48550/arXiv.2205.15688}}}

\section{Introduction}
When disasters, such as hurricanes, tsunamis, and earthquakes occur, people need to know the locations of affected residents, which is crucial for rescue and localization. Easily accessible remote sensing images enable us to identify building damage more accurately and to locate damaged buildings easier. However, this method was based on comparing images before and after the disaster manually by experts, which was time-consuming.

The introduction of deep learning methods gives another direction for automatic damage assessment, where models can classify building damage after training on labeled pre-and post-disaster remote sensing images \cite{rudner2019multi3net,kyrkou2020emergencynet,zheng2021building,gupta2021rescuenet,adriano2021learning}. Some scholars take the difficulty of obtaining large amounts of labeled data of disaster into consideration, and disaster types and datasets require different labeling methods. They adopted a semi-supervised or unsupervised approach, using only small amounts of labeled or/and unlabeled data for research \cite{lee2020assessing,xia2021building,ghosh2021unsupervised,li2020aligning,lin2021building}.

However, building an accurate model for new disasters is still far from enough. First, existing semi-supervised learning requires a certain amount of labeled data and does not fully address the problem of the complex labeling of disaster datasets \cite{lee2020assessing,xia2021building,ghosh2021unsupervised}. Second, current unsupervised learning generally relies on pre-training on labeled data \cite{li2020aligning,lin2021building}, which has not good enough for portability. Third, the interpretability of deep learning is poor, so it cannot be reliably used in our daily life.

To solve these problems, we generate supervision from entirely unlabeled datasets to clarify the features in a self-supervised setting. Also, it turns out that the new self-supervised learning using entirely unlabeled datasets can achieve similar results as supervised learning \cite{grill2020bootstrap,dosovitskiy2020image}. In the pre-training stage, a dataset containing images before and after the disaster is used, and both types of images are passed through the model to show the implied features. The model has two main parts, using contrast learning and image reconstruction which learn global information as well as local information, respectively. In the second stage, we use the weights of the pre-trained model to pass the images before and after the disaster through our model, and then join them together and pass them through a semantic segmentation head to compare with the labeled images. Finally, the effect of our model is reflected by the $F1$-score.\
The main contributions of this paper are as follows.
\begin{enumerate}
    \item [1]{We propose a novel model for building damage assessment. The model utilizes the transformer encoder structure internally, which has an excellent performance in remote sensing images. This structure is applied to build damage assessment and has achieved ideal results.}
    \item [2]{The proposed model does not require labeled data. It can learn the features of images from unlabeled data only using a contrast learning self-supervised approach. Without labeled data, our result is very close to labels.}
\end{enumerate}

\section{Related Work}
\textbf{Deep learning achievements in building damage recognition.} Gupta et al. \cite{gupta2019xbd} provided a labeled dataset including nineteen different events and more than 20,000 images, which is one of the largest and highest-quality datasets. Supervised methods have yielded excellent results with this boost \cite{rudner2019multi3net,kyrkou2020emergencynet,zheng2021building,gupta2021rescuenet,adriano2021learning}.

Supervised methods require labeled images, but the disaster domain has less labeled data than the traditional application domain of deep learning. In addition, supervised learning relies on manual labeling for training, which often leads to costly models. Lee et al. \cite{lee2020assessing} used two semi-supervised methods, MixMatch and FixMatch, to train models by fusing features from labeled and unlabeled data (containing damaged and undamaged areas), obtaining good results. Later, Xia et al. \cite{xia2021building} used less labeled data and trained the model with only a tiny number of positive examples (images of damaged areas) and the rest unlabeled, achieving results comparable to supervised learning by combining positive examples and unlabeled data. Ghosh et al. \cite{ghosh2021unsupervised} employed a novel two-part graph neural network-based framework with only a small amount of labeled data.\\
\indent However, semi-supervised methods do not fully address the problem of complex labeling; moreover, most of them have demanding requirements on labeled data. Therefore, some scholars have averted their eyes to unsupervised methods. Li et al. \cite{li2020aligning} used post-hurricane disaster data and transfer learning. They first completed pre-training on the labeled source domain, aligning the features in the source and target domains by maximum mean discrepancy (MMD), and later transferred to two new types of hurricane data to complete the classification. They \cite{lin2021building} later proposed a new generative adversarial network to align the source and target domains in an unsupervised approach to achieve better results on both transferred tasks.\\
\indent The unsupervised methods above require labeled data for pre-training and later transfer to new tasks, making the model more complex and possibly requiring repeated pre-training for different tasks, which is time-consuming and labor-intensive. Akiva et al. \cite{akiva2021h2o} used a self-supervised approach to identify flood trajectories, and their method outperformed current semantic segmentation methods for the flood trajectory segmentation task, demonstrating that self-supervised learning can perform well in disaster identification.\\
\indent In addition, Chowdhury et al. \cite{chowdhury2020comprehensive} proposed a high-resolution UAV dataset HRUD, incorporating images after Hurricane Michael, on which they wished to evaluate the performance of semantic segmentation models. They \cite{chowdhury2021attention} later gave ReDNet, which used a self-attention mechanism to achieve high accuracy on HRUD, unveiling the potential of the self-attention mechanism.

\noindent \textbf{Development of self-supervised learning.} Self-supervised learning is a category of machine learning that requires no labeled data or any pre-training on labeled data to learn features. An early self-supervised learning training model, by making features as close as possible between positive samples and as far as possible between negative samples \cite{chen2020simple,he2020momentum}, is known as contrastive learning. However, such methods need to compare features between a large number of images and often require a large batch\_size \cite{chen2020simple} or memory\_bank \cite{he2020momentum}. The BYOL \cite{grill2020bootstrap} model proposed later shows that self-supervised learning can work without differentiating the classes of images, i.e., without negative samples, but the model performance has decreased. The recently proposed DINO \cite{caron2021emerging} model uses the Transformer encoder and BYOL-based structure and solves the problem of model instability.\\
\indent Above research shows that self-supervised learning without any labeled data or pre-training can perform no less well than supervised learning.

\noindent \textbf{Transformer structure.} The Transformer \cite{vaswani2017attention} structure used a self-attention mechanism and was first proposed by the Google team in 2017, bringing significant results and breakthroughs in NLP. Inspired by the Transformer structure, the Google team proposed Vision Transformer \cite{dosovitskiy2020image} and achieved better results on image datasets. The model of MoCo v3 \cite{chen2021empirical} was the first to combine the ViT structure with self-supervised contrast learning. However, the instability generated by the combination of the two affected the results to a certain extent. DINO \cite{caron2021emerging} applied a method to solve the instability problem.\\
\indent The Transformer structure exceeds other methods in all areas and shows promising results even when confronted with instability problems by combining with contrast learning.

Inspired by the research mentioned, we conducted this study on a dataset containing pre- and post-disaster images. For the first time in the field, we employed a contrastive learning approach that combines a self-supervised method with a Transformer structure, yielding positive results.
\section{Data Description}
\begin{figure}[ht]
\centering
\includegraphics[width=0.8\textwidth]{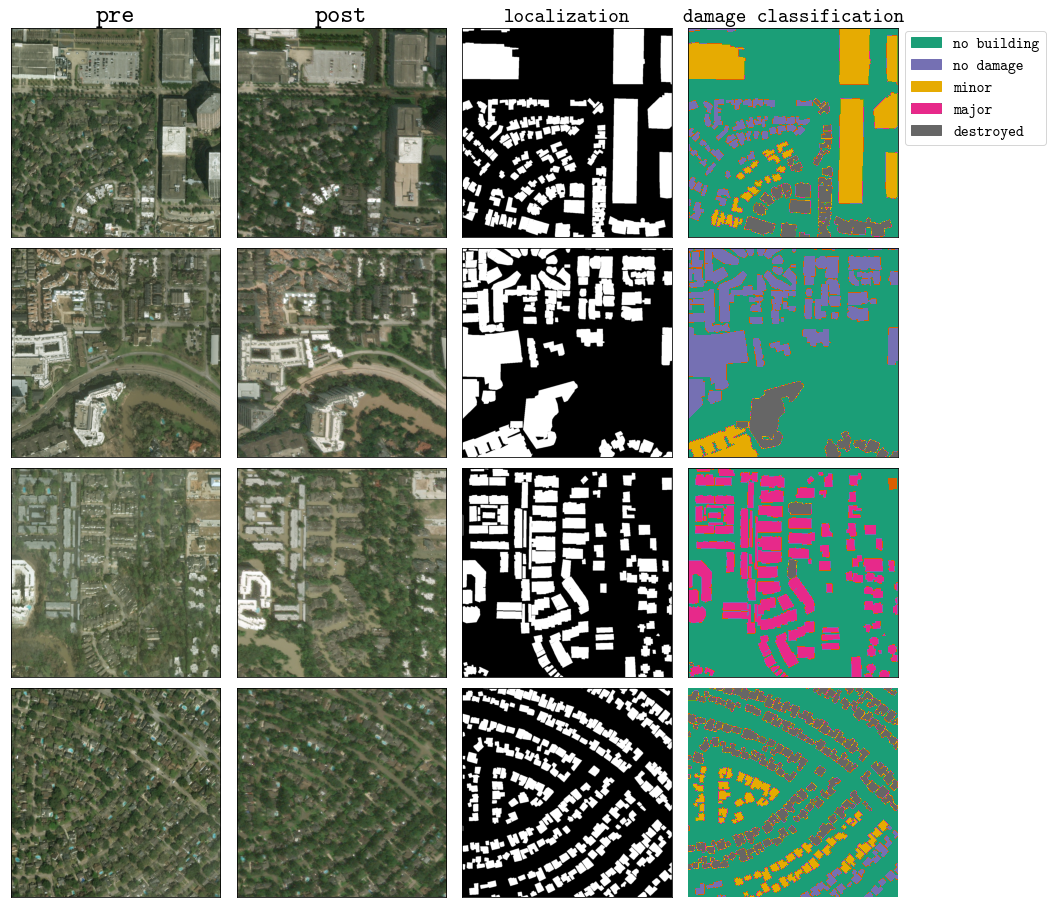}
\caption{Examples of xBD dataset}
\label{xbd}  
\end{figure}
Inspired by the literature mentioned above, we conducted our study on a dataset containing pre-and post-disaster images. For the first time in the field, we employ a contrastive learning approach that combines a self-supervised method with a Transformer structure. Combining these methods, our model yields positive results.

We conduct experiments on the xBD dataset, the largest and highest-quality satellite remote sensing image dataset for natural disasters. It contains 22,068 remote sensing images of 19 different disaster types, such as earthquakes, floods, wildfires, volcanic eruptions, and car accidents. Since there are pre-and post-disaster remote sensing images, they can be used to construct the tasks of localization and damage classification. The dataset comes from a 5,000 $km^2$ area in 15 countries with high-resolution images. Two annotations are provided for a pair of images(pre and post), respectively, for localization problem and damage classification, as shown in the Fig. \ref{xbd}.
\section{Model}
\subsection{Overview}
\begin{figure}[htbp]
\centering
\includegraphics[width=0.8\textwidth]{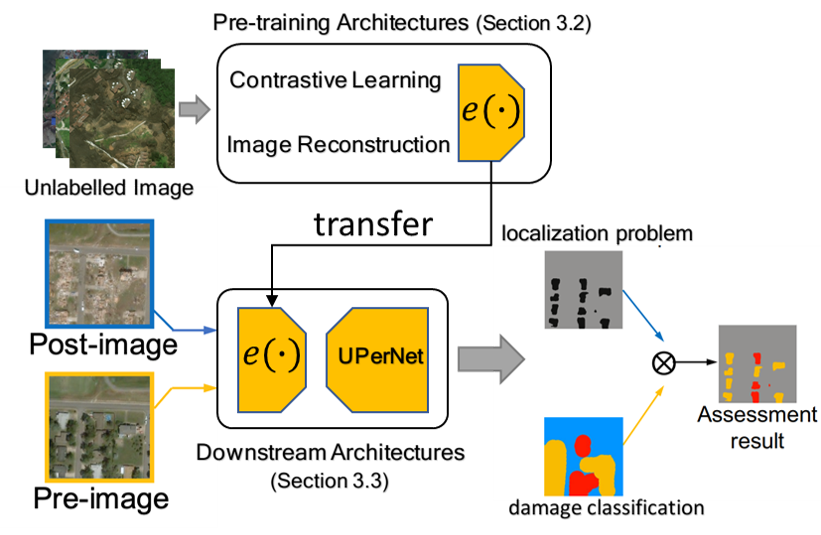}
\caption{Overview of the proposed model. The first stage utilizes pre-training architecture, while the second stage utilizes downstream architecture. Localization problem and damage classification are solved uniformly in the second stage.}
\label{zong}  
\vspace{-0.8cm}
\setlength{\abovecaptionskip}{-0.2cm}  
\setlength{\belowcaptionskip}{-1cm}  
\end{figure}
\textbf{Problem definition:} Building damage classification usually consists of two subtasks: localization and damage classification. For the former one, we classified each pixel in the pre-disaster image as “building” or “no building”. When it comes to damage classification, for the corresponding pixels in the post-disaster image, we assign values from 0 to 4 depending on the damage degree, where 0 means “no building”, 1 as “no damage”, 2 as “minor damage”, 3 as “major damage”, and 4 as “being destroyed”.

The whole process consists of two main stages. In the first stage or pre-training stage, we pre-train encoder $e(\cdot)$ on a large amount of unlabeled data using a self-supervised contrastive learning approach. Then we transfer the pre-training encoder $e(\cdot)$ to the downstream task for building damage classification (localization problem and damage classification) during the second stage.

The two stages are summarized in detail in the Fig. \ref{zong}. Localization problem and damage classification are solved uniformly in the second stage. We show the architecture of the two stages below in detail.

\subsection{Pre-training Architecture}
\begin{figure}[htbp]
\centering
\includegraphics[width=0.8\textwidth]{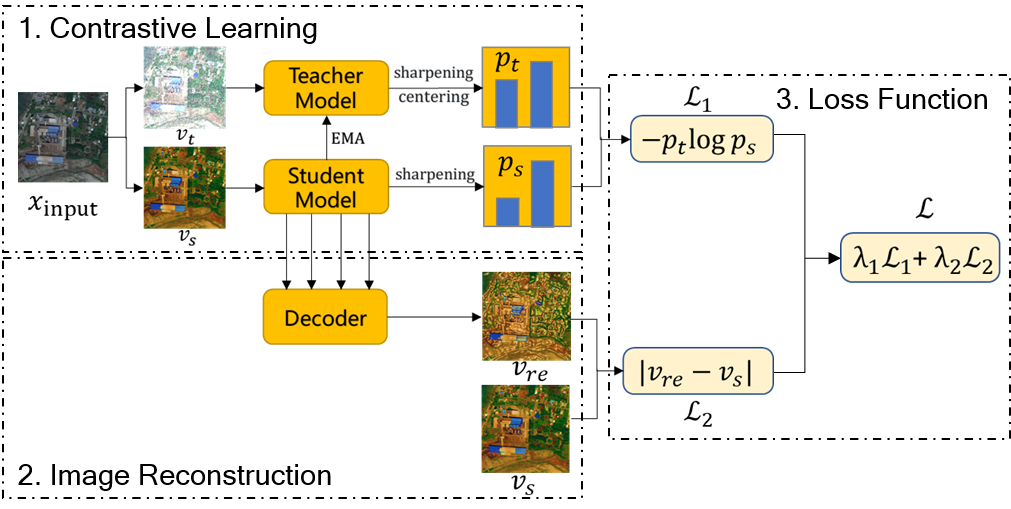}
\caption{Pre-training Architecture. Contrastive learning is displayed in the upper part, while image reconstruction is displayed in the lower part.}
\label{overview}  
\end{figure}

A self-supervised contrastive learning framework was used to pre-train encoder $e(\cdot)$ \cite{caron2021emerging} a novel style with no negative samples (DINO), which achieved good performance in transfer learning.
However, DINO was designed based on image classification and could not learn the location information of remote sensing images well. With reference to \cite{wang2022repre}, we improved Dino so that it can be suitable for remote sensing images. Fig. \ref{overview} shows an overview of our pre-training architecture. There are two parallel tasks, namely, contrastive learning and image reconstruction, which are used to learn global and local semantics.

\begin{algorithm} 
	\caption{Pre-training Architecture} 
	\label{alg3} 
	\begin{algorithmic}[1]
		\REQUIRE A set of images $\bm{D}$; student model $g_{s}(\cdot)$, teacher model $g_{t}(\cdot)$, reconstruction decoder $d(\cdot)$; parameters $\tau$, $\bm{C}$, $\lambda_1,\lambda_2$
	\FOR {sampled batch $\bm{x} \gets \bm{D}$ }
    \STATE Draw augmentations: $\bm{x_s} \gets {\rm aug1}(\bm{x}), x_t \gets {\rm aug2}(\bm{x})$
    \STATE \textbf{Contrastive learning}:
    \STATE Get output: $\bm{out_s} \gets g_{s}(\bm{x_s}), \bm{out_t} \gets g_{t}(\bm{x_t})$
    \STATE Sharpening and centering: $\bm{p_s} \gets {\rm softmax} (\bm{out_s}/\bm{\tau}), \bm{p_t} \gets {\rm softmax}(\bm{out_t}-\bm{C})$
    \STATE \textbf{Reconstruction}:
    \STATE Get multi-hierarchy features:$\{\bm{f}\} \gets e_{s}(\bm{x_s})$
    \STATE Image reconstruction:$\bm{x_{re}} \gets d(\{\bm{f}\})$
    \STATE Compute contrastive learning loss and image reconstruction loss using Eq. \ref{L1} and Eq. \ref{L2}:
    ${\mathcal L}_1 \gets CE(\bm{p_s}, \bm{p_t}), {\mathcal L}_2 \gets L1(\bm{x_{re}}, \bm{x_s})$
    \STATE Compute total loss: ${\mathcal L} \gets {\lambda}_1\mathcal{L}_1 + {\lambda}_2\mathcal{L}_2$
    \STATE Update student model $g_{s}(\cdot),\lambda_1,\lambda_2$ to minimize ${\mathcal L}$
    \STATE Update teacher model $g_{t}(\cdot)$ and $\bm{C}$ by EMA
\ENDFOR
	\end{algorithmic} 
\end{algorithm}

\subsubsection{Contrastive Learning}
\

Contrastive learning is used to learn the global semantics of images, which is instrumental for coarse-grain task. Asymmetric twin network architecture was used to learn two neural networks simultaneously, called the teacher network $g_t(\cdot)$ and the student network $g_s(\cdot)$. The network architecture is shown in the upper part of the Fig. \ref{overview}. It is widely used in contrastive learning \cite{grill2020bootstrap,caron2021emerging}. Both the teacher $g_t(\cdot)$ and the student network $g_s(\cdot)$ consists of transformer encoder $e(\cdot)$ and projector, both of which are initialized with same given weights.

Suppose the set of images is labeled as $D$. An image $x \in D$ is randomly selected, with different image augmentation transforms ${\rm aug1, aug2}$ to result in $v_s$ and $v_t$ which are different views of $x$. We input $v_s$ and $v_t$ into the student network $g_s(\cdot)$ and teacher network $g_t(\cdot)$ respectively, and the two outputs $out_s$ and $out_t$ are returned via encoder $e(\cdot)$ and projector. The corresponding probability distributions $p_s$ and $p_t$ are obtained by applying a softmax function to the outputs. The views of any randomly augmented version of a sample image should have similar feature representations. The cross-entropy loss between $p_s$ and $p_t$ is calculated for updating the parameters of the student network $\theta_s$.

\begin{figure}[htbp]
\vspace{-0.8cm}
\centering
\includegraphics[width=\textwidth]{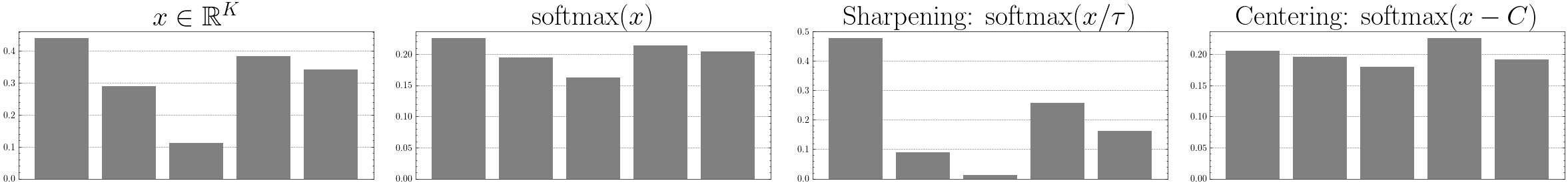}
\caption{Example of sharpening and centering. A simulated feature space has been used for an intuitive representation in low dimensions.}
\label{sharpening_and_centering}  
\vspace{-0.8cm}
\setlength{\abovecaptionskip}{-0.2cm}  
\setlength{\belowcaptionskip}{-1cm}  
\end{figure}

The asymmetry of twin networks is mainly presented in the following two points.
\begin{itemize}
\item \textbf{The two networks utilize different parameter updating methods.} The parameters of the student network $\theta_s$ were updated by minimizing the loss function $\mathcal{L}$, while those of the teacher network $\theta_t$ by exponential moving average (EMA). Specifically, the teacher network parameter $\theta_t$ update rule is ${\theta}_t \gets \lambda{\theta}_t+(1-\lambda)\theta_s$, where the hyperparameter $\lambda$ follows the cosine schedule. Referring to \cite{caron2021emerging}, such a momentum encoder update can result in better convergence of the model.
\item \textbf{Sharpening and centering.} To prevent the model from collapsing, two operations are added to the model output, sharpening and centering. However, only sharpening is applied to the student network $g_s(\cdot)$ and both operations are applied to the teacher network $g_t(\cdot)$. Both operations take place for the softmax function. Sharpening is achieved by adding a temperature parameter $\tau$ to the softmax function, as shown in Eq. \ref{sharpening}. 

\begin{equation}
P(x)=\frac{\exp \left(g_{\theta}(x) / \tau\right)}{\sum_{k=1}^{K} \exp \left(g_{\theta}(x)^{(k)} / \tau \right)}
\label{sharpening}
\end{equation}
Sharpening enhances the variance of the softmax output. As for centering, it is only applied to the teacher network where the output is made more average by subtracting the vector $C \in {\mathbb{R}}^{\rm out}$. By balancing centering with sharpening, collapse can be avoided while ensuring model convergence. Fig. \ref{sharpening_and_centering} shows the feature space after centering and sharpening. 
\end{itemize}

\subsubsection{Image Reconstruction}
\

Compared with natural images, remote sensing images variate in scale and are small in geographical size. Image reconstruction is introduced to learn better the local semantics in remote sensing images concerning \cite{wang2022repre}. As shown in the lower part of Fig. \ref{overview}, the reconstruction of the image is carried out by a lightweight decoder $d(\cdot)$ using the multi-layer feature maps $\{f\}$ output by encoder $e(\cdot)$. Since the student network $g_s(\cdot)$ retains the gradient backpropagation, the encoder of the student network $e_s(\cdot)$ is used in this task. For details, the input 3-channel RGB images $x$ are fed to the encoder of the student network $e_s(\cdot)$, and the output multi-level sequence features are fed to a lightweight decoder $d(\cdot)$. The lightweight decoder $d(\cdot)$ learns to recover the original image $x$ from the multi-layer feature maps $\{f\}$ by multi-stage feature fusion.
\begin{itemize}
    \item \textbf{Multi-layer feature maps $\{f\}$.} The output of each stage of Swin Transformer is selected as multi-layer feature maps $\{f\}$. Swin Transformer\cite{swin} is designed with reference to the layered feature representation of convolutional neural networks, where the whole model is grouped into different layers, with each downsampling the feature maps output from the previous layer, in which the layer features are calculated by moving windows. Therefore, multi-layer and multi-scale, the feature maps output by each stage of Swin Transformer have different resolutions.
    \item \textbf{Lightweight decoder $d(\cdot)$.} A lightweight decoder $d(\cdot)$ is used to recover the original image $x$ from multi-layer feature maps $\{f\}$, inspired by \cite{wang2022repre}. It is necessary to ensure that the decoder $d(\cdot)$ is lightweight, so that not only the computation can be reduced, but also the encoder $e(\cdot)$ can be better trained. The decoder $d(\cdot)$ we use consists of several fusion layers, each containing a 3 × 3 convolutional layer and a ReLU layer. The number of fusion layers corresponds to the number of feature maps, and the final output is restored to the same resolution as the original image $x$ by a 1 × 1 convolutional layer.
\end{itemize}
\subsubsection{Loss Function $\mathcal{L}$}

\

Corresponding to contrastive learning and image reconstruction, our total loss function $\mathcal{L}$ consists of two parts. For contrastive learning, the loss of the probability distribution of the student network output $p_s$ and of the teacher network output $p_t$ is calculated by cross-entropy loss (Eq. \ref{L1}).
\begin{equation}
    \mathcal{L}_1(p_s, p_t) := -p_t \log p_s
\label{L1}
\end{equation} As for the image reconstruction, $L1$ loss function is used to calculate the loss of the original image $x$ versus the decoder output $x_{\rm re}$, as shown as Eq. \ref{L2}.

\begin{equation}
    \mathcal{L}_2(x, x_{\rm re}) := \left| x-x_{\rm re} \right|
\label{L2}
\end{equation} Since our loss are applied to multiple tasks, we add learnable weights $\lambda_1, \lambda_2$ to the final loss function $\mathcal{L}$, which could be described as $\mathcal{L} := \lambda_1\mathcal{L}_1+\lambda_2\mathcal{L}_2$, where $\lambda_1, \lambda_2$ are the learnable parameter and updated with gradient descent.

\subsection{Architectures for Downstream Tasks}
\begin{figure}[htbp]
\centering
\vspace{-0.8cm}
\includegraphics[width=0.8\textwidth]{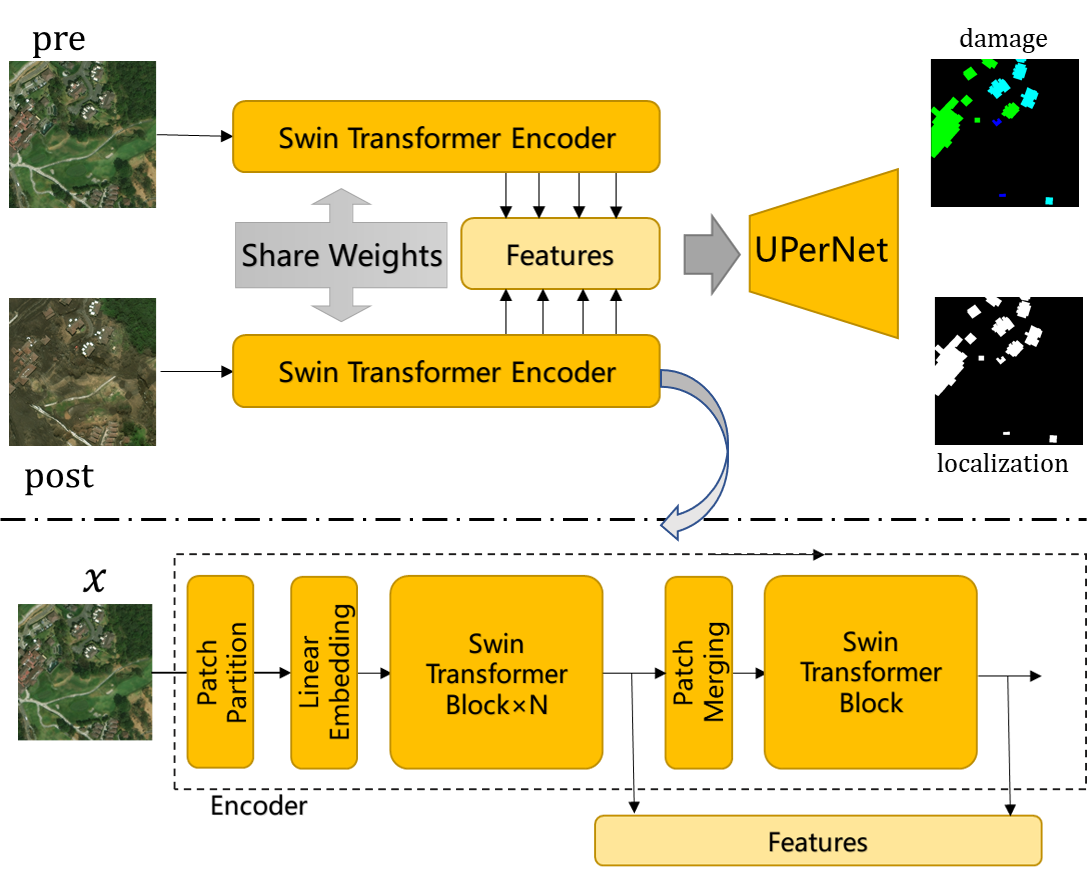}
\caption{Architectures for Downstream Tasks. This is an end-to-end model, using the common encoder-decoder architecture. The internal structure of the encoder is shown below the dotted line. Swin Transformer encoder is used.}
\label{downstream}  
\vspace{-0.8cm}
\setlength{\abovecaptionskip}{-0.2cm}  
\setlength{\belowcaptionskip}{-1cm}  
\end{figure}
In the second stage, our goal is to fine-tune the encoder, which has been pre-trained in the first stage, to be applied for building damage classification. For the downstream task, we use the Swin Transformer\cite{swin} pre-trained by contrastive learning as an encoder, with a semantic segmentation head, to handle localization and damage classification consistently, as shown in Fig. \ref{downstream}. Fig. \ref{downstream} also shows feature maps output from inside of the encoder. Regarding the multi-temporal characteristics of disaster images, our model is a siamese network. Pre- and post-disaster images are fed into two Swin Transformers, which share the weights. The output of each layer of the Swin Transformer is a 2D feature map. We concatenate the 2D feature maps corresponding to the pre-and post-disaster images along the feature dimension. These 2D feature maps are fed into the UPerNet\cite{upernet} semantic segmentation head, which undergoes multi-layer feature fusion and upsampling to finally obtain a prediction map of the original image resolution size. There are 5 channels in the feature map, and the elements of different channels represent the probability that the corresponding pixel point is predicted to be of that class. For damage classification, the prediction map is applied to the argmax function along the channel dimension to obtain a 1-channel mask. The value of each pixel in the mask is the category with the highest probability in the prediction map. For localization, referring to \cite{gupta2019xbd}, we classify the points with predicted values greater than or equal to 1 as buildings. In this way, we apply the model to both the localization and damage classification.

\section{Experiments} 
\subsection{Evaluation Metrics}
The effectiveness of self-supervised pre-training is evaluated by the performance of the model in downstream tasks and the method used in the xView2 competition to evaluate disaster damage classification is also employed in this research. In detail, $F1$-score values are calculated separately for localization and damage classification, and $F1$-score values for damage classification show the performance in different categories of disasters. Due to the unevenness of the categories, the $F1$-score can evaluate the model performance better than accuracy.
\subsection{Experiment Results}
In this section, we made comparisons with other methods to evaluate the performance of the overall model on downstream segmentation tasks. With the results of attention matrix visualization, we show the potential of the pre-training stage.
\subsubsection{Comparison with other methods.}  To evaluate the effectiveness of self-supervised contrast learning pre-training, we compared it with other pre-training methods (DINO \cite{caron2021emerging}, MoCo v3 \cite{chen2021empirical} ). Pre-training with ImageNet is a currently widespread initialization method, which was used for comparison. The entire training data is used for self-supervised pre-training.

\begin{itemize}
    \item \textbf{$F1$-score under limited annotation.} For fine-tuning on the downstream task (the second stage), we controlled the amount of labeled data and fine-tuned using 1\% and 20\% of the labeled data, respectively. We used different initialization methods with the same downstream task architecture, controlled learning rate of 1e-4, and loss function of Dice-loss with CE-loss. The results are shown in Table \ref{Tab.1}. Our method has better performance in both cases.
\begin{table}
\centering
\begin{tabular}{cccccccc}
\toprule
Amount&Method&localization&damage&No damage&Minor&Major&Destroyed\\
\midrule
\multirow{3}{*}{1\%}&ImageNet&0.461&0.321&0.387&0.136&0.234&NaN\\
\cline{2-8}
&DINO&0.522&0.366&0.480&0.157&0.439&NaN\\
\cline{2-8}
&MoCo v3&0.550&0.379&0.425&0.124&0.337&NaN\\
\cline{2-8}
&Ours&0.539&0.390&0.486&0.261&0.345&NaN\\
\midrule
\multirow{3}{*}{20\%}&ImageNet&0.661&0.587&0.604&0.278&0.471&0.456\\
\cline{2-8}
&DINO&0.714&0.601&0.667&0.229&0.384&0.447\\
\cline{2-8}
&MoCo v3&0.650&0.639&0.562&0.230&0.392&0.400\\
\cline{2-8}
&Ours&0.678&0.636&0.646&0.314&0.480&0.380\\
\bottomrule
\end{tabular}
\caption{$F1$-score under limited labeled data. Nan means that the model has no valid output for that category, which is a result of insufficient samples.}
\label{Tab.1}
\vspace{-0.8cm}
\setlength{\abovecaptionskip}{-0.2cm}  
\setlength{\belowcaptionskip}{-1cm}  
\end{table}

\item \textbf{10\% amount of annotated data with training process.}  We explored the performance of the pre-training model on limited labeled data by training it longer. Using 1\% labeled data with a learning rate of 1e-4, we trained more epochs while fine-tuning (the second stage). Using our pre-training model, and ImageNet pre-training model with random initialization method respectively, 500 epochs have been tested to compare the training process. The results are shown in Fig. \ref{result_loss_val}. It turns out that our model loss converges faster and has a higher $F1$-score.
\begin{figure}[htbp]
\vspace{-0.8cm}   %调整图片与上文的垂直距离 
\setlength{\belowdisplayskip}{3pt} 	
\centering
\includegraphics[width=\textwidth]{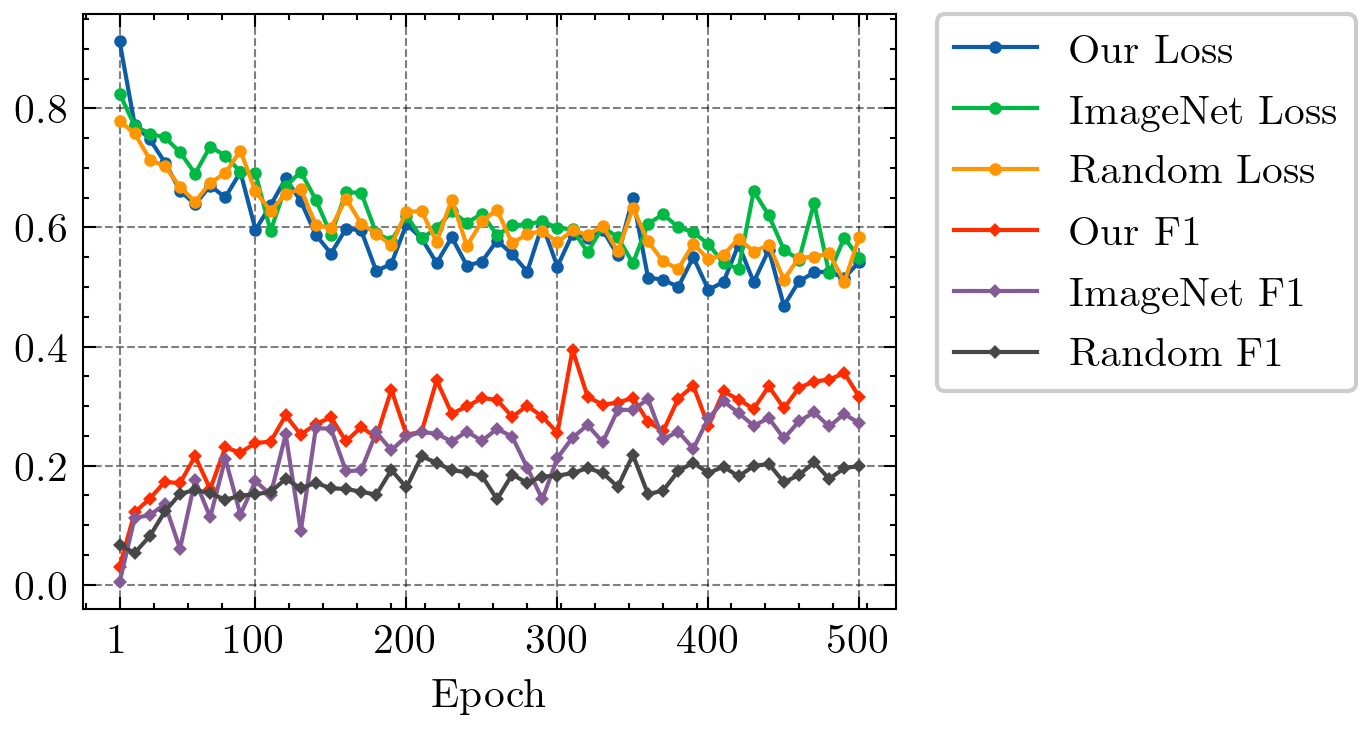}
\caption{10\% labeled data with training process. The line graph was eventually plotted as we recorded every 5 epochs.}
\label{result_loss_val}  
\vspace{-0.8cm}
\setlength{\abovecaptionskip}{-0.2cm}  
\setlength{\belowcaptionskip}{-1cm}  
\end{figure}
\end{itemize}
\subsubsection{Results of attention matrix visualization.} 
\begin{figure}[h]
\centering
\includegraphics[width=0.8\textwidth]{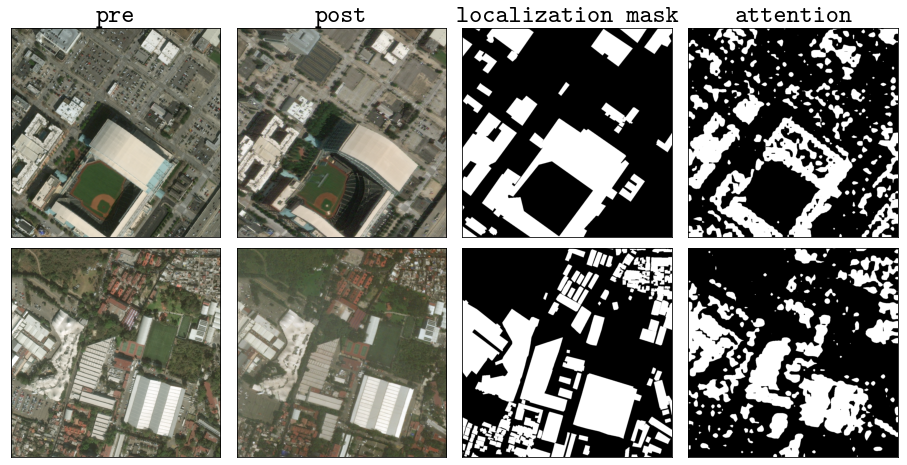}
\caption{Pre, post, mask, and attention matrix. The result is obtained from the first stage(pre-training), and we don't use any annotations in the first stage. Attention matrix, which was trained without labeled data, is very close to the man-made mask.}
\label{attention}  
\vspace{-0.8cm}  %调整图片与上文的垂直距离
\setlength{\abovecaptionskip}{-0.2cm}  
\setlength{\belowcaptionskip}{-1cm}  
\end{figure}

We visualized the attention matrix of encoder in the pre-training stage. Though the pre-training does not have labeled data involved, the visualization results of the attention matrix are rich in semantic information. Fig. \ref{attention} shows the visualization results of the attention matrix in the transformer, where our model learns the approximation without any labeled information.
\setlength{\parskip}{0.2cm plus4mm minus3mm}
\section{Conclusion}
In this paper, we design a new self-supervised deep learning method to assess the damage level of buildings with satellite images before and after a disaster. In the first stage, we use self-supervised learning to learn feature representations from unlabeled data and achieve results close to mask manual labeled. As for the second stage, we use the model of the first stage as an encoder to splice the pre- and post-disaster features before and after a semantic segmentation network to obtain the disaster-determined images for building identification as well as disaster assessment.

An important contribution of this study is the use of a self-supervised approach in building damage assessment, which enables the model to be trained with less reliance on labeled data. We also solved the difficulties caused by inconsistent satellite image labeling and disaster types in this field in the past. We can use only the original remote sensing images before and after the disaster to generate the disaster damage images of an area and assess the damage level of buildings.  The results show an accuracy rate with the contrast learning approach that is significantly better than the existing baseline, and the visualization results in the unlabeled case are close to the manual labeling. The self-supervised approach has been proved to have outperforming results and considerable potential for building damage assessment. In the future, we plan to explore the combination of self-supervised models deeper with building damage identification using multi-temporal information and improve the model architecture for better performance.

\noindent \textbf{Acknowledgements.} This research was supported by Public Health \& Disease Control and Prevention, Major Innovation \& Planning Interdisciplinary Platform for the “Double-First Class” Initiative, Renmin University of China” (No. 2022PDPC), fund for building world-class universities (disciplines) of Renmin University of China. Project No. KYGJA2022001, fund for building world-class universities (disciplines) of Renmin University of China. Project No. KYGJF2021001, Beijing Golden Bridge Project seed fund (No. ZZ21021). This research was supported by Public Computing Cloud, Renmin University of China. 


\begin{thebibliography}{10}
% \providecommand{\url}[1]{#1}
% \csname url@samestyle\endcsname
% \providecommand{\newblock}{\relax}
% \providecommand{\bibinfo}[2]{#2}
% \providecommand{\BIBentrySTDinterwordspacing}{\spaceskip=0pt\relax}
% \providecommand{\BIBentryALTinterwordstretchfactor}{4}
% \providecommand{\BIBentryALTinterwordspacing}{\spaceskip=\fontdimen2\font plus
% \BIBentryALTinterwordstretchfactor\fontdimen3\font minus
%   \fontdimen4\font\relax}
% \providecommand{\BIBforeignlanguage}[2]{{%
% \expandafter\ifx\csname l@#1\endcsname\relax
% \typeout{** WARNING: IEEEtran.bst: No hyphenation pattern has been}%
% \typeout{** loaded for the language `#1'. Using the pattern for}%
% \typeout{** the default language instead.}%
% \else
% \language=\csname l@#1\endcsname
% \fi
% #2}}
% \providecommand{\BIBdecl}{\relax}
% \BIBdecl

\bibitem{rudner2019multi3net}
T.~G. Rudner, M.~Ru{\ss}wurm, J.~Fil, R.~Pelich, B.~Bischke,
  V.~Kopa{\v{c}}kov{\'a}, and P.~Bili{\'n}ski, ``Multi3net: segmenting flooded
  buildings via fusion of multiresolution, multisensor, and multitemporal
  satellite imagery,'' in \emph{Proceedings of the AAAI Conference on
  Artificial Intelligence}, vol.~33, no.~01, 2019, pp. 702--709.

\bibitem{kyrkou2020emergencynet}
C.~Kyrkou and T.~Theocharides, ``Emergencynet: Efficient aerial image
  classification for drone-based emergency monitoring using atrous
  convolutional feature fusion,'' \emph{IEEE Journal of Selected Topics in
  Applied Earth Observations and Remote Sensing}, vol.~13, pp. 1687--1699,
  2020.

\bibitem{zheng2021building}
Z.~Zheng, Y.~Zhong, J.~Wang, A.~Ma, and L.~Zhang, ``Building damage assessment
  for rapid disaster response with a deep object-based semantic change
  detection framework: From natural disasters to man-made disasters,''
  \emph{Remote Sensing of Environment}, vol. 265, p. 112636, 2021.

\bibitem{gupta2021rescuenet}
R.~Gupta and M.~Shah, ``Rescuenet: Joint building segmentation and damage
  assessment from satellite imagery,'' in \emph{2020 25th International
  Conference on Pattern Recognition (ICPR)}.\hskip 1em plus 0.5em minus
  0.4em\relax IEEE, 2021, pp. 4405--4411.

\bibitem{adriano2021learning}
B.~Adriano, N.~Yokoya, J.~Xia, H.~Miura, W.~Liu, M.~Matsuoka, and S.~Koshimura,
  ``Learning from multimodal and multitemporal earth observation data for
  building damage mapping,'' \emph{ISPRS Journal of Photogrammetry and Remote
  Sensing}, vol. 175, pp. 132--143, 2021.

\bibitem{lee2020assessing}
J.~Lee, J.~Z. Xu, K.~Sohn, W.~Lu, D.~Berthelot, I.~Gur, P.~Khaitan,
  K.~Koupparis, B.~Kowatsch \emph{et~al.}, ``Assessing post-disaster damage
  from satellite imagery using semi-supervised learning techniques,''
  \emph{arXiv preprint arXiv:2011.14004}, 2020.

\bibitem{xia2021building}
J.~Xia, N.~Yokoya, and B.~Adriano, ``Building damage mapping with
  self-positiveunlabeled learning,'' \emph{arXiv preprint arXiv:2111.02586},
  2021.

\bibitem{ghosh2021unsupervised}
S.~Ghosh, S.~Maji, and M.~S. Desarkar, ``Unsupervised domain adaptation with
  global and local graph neural networks in limited labeled data scenario:
  Application to disaster management,'' \emph{arXiv preprint arXiv:2104.01436},
  2021.

\bibitem{li2020aligning}
Y.~Li, W.~Hu, H.~Li, H.~Dong, B.~Zhang, and Q.~Tian, ``Aligning discriminative
  and representative features: An unsupervised domain adaptation method for
  building damage assessment,'' \emph{IEEE Transactions on Image Processing},
  vol.~29, pp. 6110--6122, 2020.

\bibitem{lin2021building}
C.~Lin, Y.~Li, Y.~Liu, X.~Wang, and S.~Geng, ``Building damage assessment from
  post-hurricane imageries using unsupervised domain adaptation with enhanced
  feature discrimination,'' \emph{IEEE Transactions on Geoscience and Remote
  Sensing}, vol.~60, pp. 1--10, 2021.

\bibitem{gupta2019xbd}
R.~Gupta, R.~Hosfelt, S.~Sajeev, N.~Patel, B.~Goodman, J.~Doshi, E.~Heim,
  H.~Choset, and M.~Gaston, ``xbd: A dataset for assessing building damage from
  satellite imagery,'' \emph{arXiv preprint arXiv:1911.09296}, 2019.

\bibitem{chowdhury2020comprehensive}
T.~Chowdhury, M.~Rahnemoonfar, R.~Murphy, and O.~Fernandes, ``Comprehensive
  semantic segmentation on high resolution uav imagery for natural disaster
  damage assessment,'' in \emph{2020 IEEE International Conference on Big Data
  (Big Data)}.\hskip 1em plus 0.5em minus 0.4em\relax IEEE, 2020, pp.
  3904--3913.

\bibitem{chowdhury2021attention}
T.~Chowdhury and M.~Rahnemoonfar, ``Attention based semantic segmentation on
  uav dataset for natural disaster damage assessment,'' in \emph{2021 IEEE
  International Geoscience and Remote Sensing Symposium IGARSS}.\hskip 1em plus
  0.5em minus 0.4em\relax IEEE, 2021, pp. 2325--2328.

\bibitem{chen2020simple}
T.~Chen, S.~Kornblith, M.~Norouzi, and G.~Hinton, ``A simple framework for
  contrastive learning of visual representations,'' in \emph{International
  conference on machine learning}.\hskip 1em plus 0.5em minus 0.4em\relax PMLR,
  2020, pp. 1597--1607.

\bibitem{he2020momentum}
K.~He, H.~Fan, Y.~Wu, S.~Xie, and R.~Girshick, ``Momentum contrast for
  unsupervised visual representation learning,'' in \emph{Proceedings of the
  IEEE/CVF conference on computer vision and pattern recognition}, 2020, pp.
  9729--9738.

\bibitem{grill2020bootstrap}
J.-B. Grill, F.~Strub, F.~Altch{\'e}, C.~Tallec, P.~Richemond, E.~Buchatskaya,
  C.~Doersch, B.~Avila~Pires, Z.~Guo, M.~Gheshlaghi~Azar \emph{et~al.},
  ``Bootstrap your own latent-a new approach to self-supervised learning,''
  \emph{Advances in Neural Information Processing Systems}, vol.~33, pp.
  21\,271--21\,284, 2020.

\bibitem{caron2021emerging}
M.~Caron, H.~Touvron, I.~Misra, H.~J{\'e}gou, J.~Mairal, P.~Bojanowski, and
  A.~Joulin, ``Emerging properties in self-supervised vision transformers,'' in
  \emph{Proceedings of the IEEE/CVF International Conference on Computer
  Vision}, 2021, pp. 9650--9660.

\bibitem{vaswani2017attention}
A.~Vaswani, N.~Shazeer, N.~Parmar, J.~Uszkoreit, L.~Jones, A.~N. Gomez,
  {\L}.~Kaiser, and I.~Polosukhin, ``Attention is all you need,''
  \emph{Advances in neural information processing systems}, vol.~30, 2017.

\bibitem{dosovitskiy2020image}
A.~Dosovitskiy, L.~Beyer, A.~Kolesnikov, D.~Weissenborn, X.~Zhai,
  T.~Unterthiner, M.~Dehghani, M.~Minderer, G.~Heigold, S.~Gelly \emph{et~al.},
  ``An image is worth 16x16 words: Transformers for image recognition at
  scale,'' \emph{arXiv preprint arXiv:2010.11929}, 2020.

\bibitem{chen2021empirical}
X.~Chen, S.~Xie, and K.~He, ``An empirical study of training self-supervised
  visual transformers,'' \emph{arXiv e-prints}, pp. arXiv--2104, 2021.

\bibitem{akiva2021h2o}
P.~Akiva, M.~Purri, K.~Dana, B.~Tellman, and T.~Anderson, ``H2o-net:
  Self-supervised flood segmentation via adversarial domain adaptation and
  label refinement,'' in \emph{Proceedings of the IEEE/CVF Winter Conference on
  Applications of Computer Vision}, 2021, pp. 111--122.

\bibitem{wang2022repre}
L.~Wang, F.~Liang, Y.~Li, W.~Ouyang, H.~Zhang, and J.~Shao, ``Repre: Improving
  self-supervised vision transformer with reconstructive pre-training,''
  \emph{arXiv preprint arXiv:2201.06857}, 2022.

\bibitem{weber2020building}
E.~Weber and H.~Kan{\'e}, ``Building disaster damage assessment in satellite
  imagery with multi-temporal fusion,'' \emph{arXiv preprint arXiv:2004.05525},
  2020.

\bibitem{swin}
Z.~Liu, Y.~Lin, Y.~Cao, H.~Hu, Y.~Wei, Z.~Zhang, S.~Lin, and B.~Guo, ``Swin
  transformer: Hierarchical vision transformer using shifted windows,'' in
  \emph{Proceedings of the IEEE/CVF International Conference on Computer
  Vision}, 2021, pp. 10\,012--10\,022.

\bibitem{upernet}
T.~Xiao, Y.~Liu, B.~Zhou, Y.~Jiang, and J.~Sun, ``Unified perceptual parsing
  for scene understanding,'' in \emph{Proceedings of the European Conference on
  Computer Vision (ECCV)}, 2018, pp. 418--434.
\end{thebibliography}
\end{document}